\pdfoutput=1

\documentclass[11pt]{article}

\usepackage[final]{coling}

\usepackage{times}
\usepackage{latexsym}

\usepackage[T1]{fontenc}

\usepackage[utf8]{inputenc}

\usepackage{microtype}

\usepackage{inconsolata}

\usepackage{graphicx}
\usepackage{subfigure}
\usepackage{multirow}
\usepackage{amsmath}
\usepackage{xcolor}
\usepackage{enumitem}
\usepackage{booktabs} 
\usepackage{siunitx} 
\usepackage{caption}

\newenvironment{customquote}%
  {\begin{quote}\itshape} 
  {\end{quote}}

\usepackage{fancyhdr}

\fancypagestyle{firstpage}{
    \fancyhf{}
    \fancyfoot[C]{Proceedings of the 31st International Conference on Computational Linguistics January 19--24, 2025. \copyright 2025 Association for Computational Linguistics}
    \renewcommand{\headrulewidth}{0pt}
}

\renewcommand{\headrulewidth}{0pt}
  
\title{A Context-Aware Approach for Enhancing Data Imputation with Pre-trained Language Models}

\author{Ahatsham Hayat  \hspace{10.00mm}   Mohammad Rashedul Hasan \\
Electrical and Computer Engineering \\
University of Nebraska-Lincoln \\
\small\texttt{aahatsham2@huskers.unl.edu, hasan@unl.edu}   \\
}

\begin{document}

\pagestyle{fancy}
\fancyhf{}
\fancyfoot[C]{\thepage}
\renewcommand{\headrulewidth}{0pt}

\maketitle

\begin{abstract}

This paper presents a novel approach named \textbf{C}ontextually \textbf{R}elevant \textbf{I}mputation leveraging pre-trained \textbf{L}anguage \textbf{M}odels (\textbf{CRILM}) for handling missing data in tabular datasets. Instead of relying on traditional numerical estimations, CRILM uses pre-trained language models (LMs) to create contextually relevant descriptors for missing values. This method aligns datasets with LMs' strengths, allowing large LMs to generate these descriptors and small LMs to be fine-tuned on the enriched datasets for enhanced downstream task performance. Our evaluations demonstrate CRILM's superior performance and robustness across MCAR, MAR, and challenging MNAR scenarios, with up to a 10\% improvement over the best-performing baselines. By mitigating biases, particularly in MNAR settings, CRILM improves downstream task performance and offers a cost-effective solution for resource-constrained environments.
\end{abstract}

\section{Introduction}
\label{intro}

\begin{customquote}
`Well! I’ve often seen a cat without a grin,' thought Alice; `but a grin without a cat! It’s the most curious thing I ever saw in all my life!' 

Lewis Carroll, Alice's Adventures in Wonderland (1865)
\end{customquote}

\vspace{0.25mm}

Missing data in tabular datasets is a ubiquitous problem often arising from real-life data collection processes \cite{kumar_2017}. Handling missing data is crucial for downstream machine learning (ML) tasks, necessitating data imputation to fill in missing entries with plausible values. However, imputation that overlooks the data context can introduce unintended biases, leading to aberrant model behavior \cite{schelter_2018, schelter_2021, garcia-laencina_2010, stoyanovich_2020, yang_2020, abedjan_2018}.

Data may be missing because it was never collected or because collected data was lost. These causes are driven by \textbf{domain-specific contexts}. For example, in the medical domain, data might not be collected due to various reasons, such as a patient's characteristics not being recorded during a visit, some tests not being performed, intentional omissions by patients, or the difficulty and danger of acquiring certain information \cite{yoon_2017_personalized_healthcare, alaa_2018, yoon_2018_cardiac_transplantation}. Data loss can occur through application or transmission errors or due to data integration errors.

Typically, imputation methods estimate missing values based on observed data, such as a patient's blood pressure and heart rate \cite{yoon_2018_deep_sensing}. However, missing data do not always depend on the observed data. Rubin's widely used categorization of missingness mechanisms identifies three cases \cite{rubin_1976}: missing completely at random (MCAR), missing at random (MAR), and missing not at random (MNAR). In MCAR, the missingness is independent of the data, whereas in MAR, the probability of being missing depends only on observed values. In MNAR, the probability of missingness depends on unobserved values, and imputation in this case can introduce significant biases to the data. \textit{Therefore, to achieve accurate imputation, it is crucial for methods to account for the specific context of the missingness}.

Existing imputation methods use various numeric estimation techniques to capture the data context, preserving joint and marginal distributions of the imputed data. Many methods, including traditional statistical approaches and machine/deep learning methods, aim to learn the joint distribution of the data either implicitly or explicitly \cite{vanBuuren_2006, yoon_2018_gain_PMLR, gondara_2018_MIDA, mattei_2019_MIWAE, nazabal_2020, zhao_2023_TDM}. However, these methods have several limitations: often requiring fully observed training data, being challenging to implement, needing separate models for each feature, and lacking support for column-specific modeling. See Section \ref{related} for more details. Moreover, most approaches, with notable exceptions \cite{kim_2018_MNAR, mohan_2019}, primarily address MCAR and MAR data, struggling with the more challenging yet prevalent MNAR case \cite{muzellec_2020}.

Parallel to numeric estimation-based imputation, we explore alternative methods for capturing data context to handle missing values. Specifically, we examine whether it is possible to bypass modeling the data distribution entirely. In scenarios where numeric-estimation methods may introduce bias, such as in MNAR settings, or prove inadequate, we develop an approach that avoids direct estimation of missing values. Instead of estimating missing values directly, we investigate whether an artificial intelligence (AI) model can implicitly handle missingness through its prior general knowledge.

To address these challenges, we explore the potential of utilizing general-purpose pre-trained language models (LMs) \cite{brown_language_2020, chowdhery_palm_2022, touvron_llama_2023, openai_gpt-4_2023} for handling diverse missingness in tabular datasets. These models possess expansive knowledge \cite{raffel_exploring_2020, roberts_how_2020}, reasoning capabilities \cite{chowdhery_palm_2022, wei_chain_thought_2023, bhatia_tart_2023}, and extensive linguistic expertise \cite{petroni_2019, mahowald_2024}, and have demonstrated exceptional performance across various downstream natural language processing (NLP) tasks \cite{bubeck_2023, raffel_exploring_2020, yang_2024_chatgpt}. Our aim is to leverage the advanced capabilities of LMs to enhance the performance of downstream tasks on tabular data with missing values.

To achieve this goal, we approach the downstream task by treating it as an NLP problem and harnessing the capabilities of LMs to handle missing values. We propose a novel method named \textbf{C}ontextually \textbf{R}elevant \textbf{I}mputation leveraging pre-trained \textbf{L}anguage \textbf{M}odels (\textbf{CRILM}), which operates through a \textit{dual-phase process}. Initially, large LMs (LLMs), such as those with more than 10 billion parameters, generate contextually relevant natural language descriptors for missing values. For instance, in the UCI Wine dataset \cite{wine_UCI_1991}, a contextually relevant descriptor for missing values in the feature malic acid could be: \textit{Malic acid quantity missing for this wine sample}. These descriptors replace missing values, transforming numeric datasets into natural language contextualized formats, thereby aligning the data with the strengths of LMs and augmenting their processing capabilities.



Subsequently, these missingness-aware textual datasets are used for solving downstream tasks such as classification, modeled as NLP tasks. The textual datasets serve as the foundation for fine-tuning \textbf{smaller pre-trained LMs} such as those with less than 10 billion parameters, showcasing a unique and effective use of language models beyond their conventional applications. By incorporating \textbf{contextually relevant descriptors for missing data}, CRILM addresses variability and specificity across different domains and navigates the complexities of various missingness mechanisms.

Recently, Transformer-based \cite{vaswani_attention_2017} methods have been proposed to handle missing values in tabular data, such as masked Transformer for generating synthetic tabular data \cite{gulati_2023_tabmt} and pre-training LMs using enriched tabular data \cite{yang_2024}. However, these approaches overlook diverse missingness patterns, raising questions about their ability to address the biases introduced by the imputation methods and whether downstream task performance improves as a result. Through the innovative integration of LMs into the data imputation process, CRILM aims to deliver a more nuanced, accurate, and reliable method for handling missing data in a context-aware fashion, essential for improving the quality of downstream NLP tasks.



Our approach offers a \textbf{cost-effective solution} by leveraging publicly available LLMs for zero-shot inference and employing smaller LMs for downstream tasks, which can be efficiently fine-tuned in low-resource environments. This feasibility is demonstrated through experiments using accessible resources like ChatGPT-3.5 \cite{openai_2023_chatgpt} for inference and smaller LM-based fine-tuning, ensuring efficient implementation.



To evaluate CRILM's effectiveness, we analyze its performance across three missing data mechanisms—MCAR, MAR, and MNAR \cite{rubin_1976}. CRILM is compared against various existing imputation methods, investigating different phrasing choices for missingness descriptors in LM-based tasks. We also explore the influence of decoder-only and encoder-decoder pre-trained LMs on downstream transfer learning, assessing their impact on task performance. Our empirical studies address two key research questions (RQs):

%

\begin{itemize}[noitemsep, topsep=2pt, left=2pt]
\item \textbf{[RQ1]}: To what extent does CRILM effectively perform in imputing missing values across distinct missingness mechanisms (MCAR, MAR, and MNAR), compared to existing methods, in terms of accuracy and robustness on varied datasets?


\item \textbf{[RQ2]}: How do feature-specific versus generic missingness descriptors impact the performance of LM-based downstream tasks?

\end{itemize}

The contributions of this work are multifaceted. \textit{Firstly}, CRILM introduces an innovative imputation approach for missing values in tabular datasets, running parallel to existing numeric-estimation-based methods. By utilizing LMs to generate context-specific descriptors for missing data, CRILM sets a new benchmark in data imputation, departing from traditional numerical methods. \textit{Secondly}, our empirical evaluation highlights CRILM's superior performance over existing methods across varied datasets and missingness patterns, particularly excelling in MNAR settings where biases introduced by numeric-estimation-based techniques can be significant. Specifically, CRILM demonstrates a substantial performance lead of up to 10\% over the best-performing baseline imputation method in the challenging MNAR scenarios. \textit{Thirdly}, we advance the understanding of the NLP capabilities of pre-trained LMs by demonstrating their potential in handling complex data imputation tasks. Additionally, the cost-effectiveness of our approach, achieved by leveraging smaller LMs for transfer learning, enhances its practicality and accessibility. \textit{Lastly}, our analysis comparing feature-specific and generic descriptors offers insights into optimizing LM performance for imputation tasks, emphasizing contextual accuracy. These contributions advance data preprocessing techniques and open novel pathways for leveraging LMs in addressing complex data science challenges.

\section{Method}
\label{method}
\subsection{Problem Formulation}

%

Consider a tabular dataset represented by a matrix $\mathbf{X}$ consisting of a collection of $n$ instances (rows) where each instance $\mathbf{X^i}$ is a \textit{d}-dimensional random variable: $\mathbf{X^i} = (X^i_1, ..., X^i_d)$ (thus $d$ columns). These variables are continuous and/or categorical. The dataset $\mathbf{X}$ has an observed portion denoted by $\mathbf{X_O}$ and a missing portion denoted by $\mathbf{X_M}$. The missingness pattern in $\mathbf{X}$ is denoted by $\mathbf{M}$, which is a matrix of the same dimensions as $\mathbf{X}$ in which cells have a value of 1 if missing and 0 otherwise. 

CRILM takes $\mathbf{X}$ and transforms it into a missingness-aware contextualized natural language dataset $\mathbf{X_{missingness\_aware}}$ by replacing the missing values by contextually relevant descriptors. Our goal is to demonstrate the efficacy of CRILM via the performance of a downstream classification task by fine-tuning an LM using $\mathbf{X_{missingness\_aware}}$.

\subsection{Generating Missing Values} 
We construct synthetic datasets with up to 30\% missing values by applying the following three missingness mechanisms on complete datasets: MCAR, MAR and MNAR. The implementations of these mechanisms are modified from~\cite{jager_2021}.

\vspace{1.00mm}
 \noindent \textbf{MCAR.} It is introduced by randomly removing 30\% of the observations from each feature.

\vspace{1.00mm}
\noindent \textbf{MAR.} First, we select all observations within the 30\textsuperscript{th} percentile range of an independent feature, typically the first column in the dataset. Then, we randomly remove 60\% of the values from each corresponding (dependent) feature within this subset, ensuring that missingness is related to the independent feature but random within the dependent features.


\vspace{1.00mm}
\noindent \textbf{MNAR.} We remove the observations of a feature if the observations fall within the 30\textsuperscript{th} percentile range of the feature value.


\subsection{Description of CRILM}
\label{CRILM}
\begin{figure}[htb!]
\centering
\includegraphics[width=0.5\textwidth]{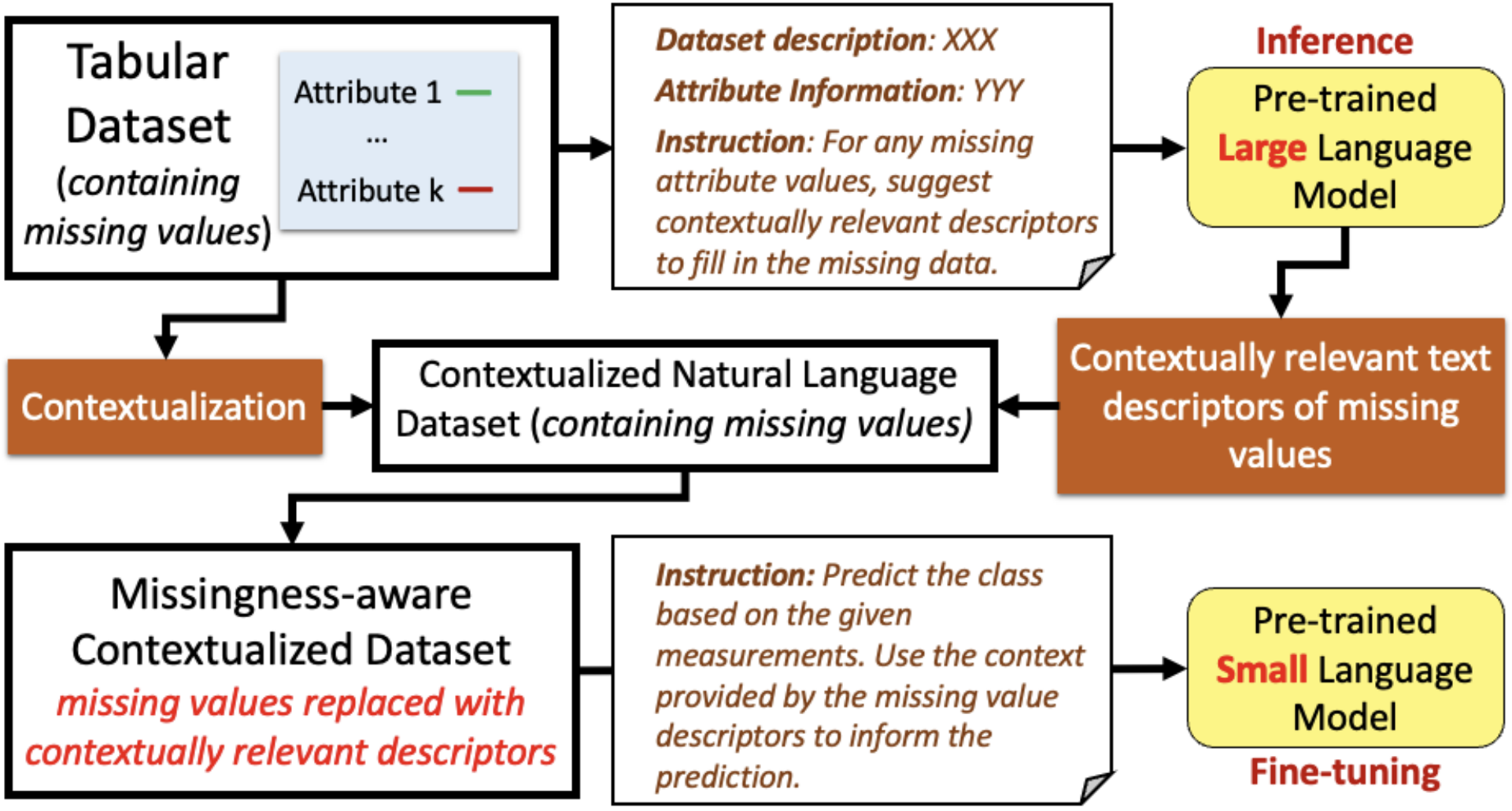}
\caption{An overview of CRILM.}
\label{fig:method}
\vspace{-5.00mm}
\end{figure}

Figure \ref{fig:method} illustrates the CRILM process, which encompasses four stages: (1) constructing a contextualized natural language dataset, (2) generating suitable descriptors for missing values, (3) creating a missingness-aware contextualized dataset, and (4) adapting an LM for downstream tasks. We detail these stages below.

\vspace{2.00mm}
\noindent \textbf{Constructing a Contextualized Natural Language Dataset.}
We construct a contextualized natural language dataset from a numeric dataset $\mathbf{X}$ containing missing values. The objective is to generate contextually suitable description of each attribute and its measures in natural language. For instance, a record from the UCI Wine dataset \cite{wine_UCI_1991} with numeric input and output attributes is contextualized as follows: \textit{``The alcohol content in the wine is 12.47. The level of malic acid in the wine is 1.52 ... The class of the wine is classified as class 1 wine.''}\footnote{The Python script used for contextualization is provided in the Supplementary Material.} This step converts numeric values into detailed descriptions, preparing the dataset for embedding missing value descriptors.


\vspace{2.00mm}
\noindent \textbf{Generating Suitable Descriptors for Missing Values.}
Unlike conventional imputation methods that estimate missing values from observed data using numerical methods, we utilize contextually relevant descriptors of missing values for imputation. We generate these descriptors by a conversational LLM (e.g., OpenAI's ChatGPT-3.5 \cite{openai_2023_chatgpt}). We prompt the LLM with a dataset description and instruct it to generate missing value descriptors, such as: \textit{``For any missing attribute values, suggest contextually relevant descriptors to fill in the missing data.}'' This method relies on LLM's extensive knowledge base and linguistic capabilities to produce appropriate missing value descriptors. A list of feature-specific contextually relevant missing-value descriptors for selected datasets are provided in Appendix \ref{appendix:descriptors}.

\vspace{2.00mm}
\noindent \textbf{Creating a Missingness-Aware Contextualized Dataset.}
We construct the missingness-aware contextualized natural language dataset, denoted as $\mathbf{X_{missingness\_aware}}$, by replacing the missing values with generated descriptors. This process ensures that each data instance is ``aware'' of its missing attributes, thereby enhancing the downstream LM's ability to learn from incomplete data by providing explicit context. Additionally, we use distinct descriptors for different features in the dataset that contain missing values. This approach implicitly informs the downstream LM to handle the missingness of each feature in a contextually appropriate manner, ultimately improving the performance of the downstream task.

\vspace{2.00mm}
\noindent \textbf{Adapting an LM for Solving Downstream Tasks.}
The final step involves fine-tuning a pre-trained small LM with the missingness-aware, contextually-rich dataset. During the fine-tuning process, we incorporate specific task instructions and strategies for handling missing data. For instance, in classification tasks, we include instructions such as: \textit{``Predict the class based on the given measurements. Use the context provided by the missing value descriptors to inform the prediction.}'' This approach ensures that an LM effectively utilizes the contextual information embedded in the descriptors, thereby enhancing its predictive performance despite the presence of missing data. Using smaller LMs for fine-tuning not only makes the process cost-effective but also allows for efficient adaptation to the specific characteristics of the dataset and task at hand.


\section{Related Work}
\label{related}

The challenge of missing data in tabular datasets has led to the development of numerous imputation methods, broadly categorized into those modeling feature distribution and those that do not. The latter category includes methods such as distribution matching and traditional non-parametric methods. In the former category, two distinct types of imputation methods exist: those treating features separately and those treating them jointly. Separate feature treatment methods, like Multivariate Imputation by Chained Equations (MICE) \cite{vanBuuren_2006, vanBuuren_2011}, which is an iterative method as well as a discriminative method, specify a univariate model for each feature based on others, with other notable iterative methods also existing \cite{heckerman_2000, raghunathan_2001, gelman_2004, liu_2014, zhu_2015}. Joint treatment methods aim to learn a joint distribution of all features, with recent developments including deep learning-based generative methods like GAIN \cite{yoon_2018_gain_PMLR}, utilizing Generative Adversarial Nets \cite{goodfellow_2014_generative}, although their effectiveness varies compared to traditional methods \cite{jager_2021}. Other types of generative models that are based on Denoising Autoencoders  \cite{vincent_2008}, have been proposed \cite{gondara_2018_MIDA, rezende_2014, mattei_2018, nazabal_2020, ivanov_2018, richardson_2020, mattei_2019_MIWAE}, though most of these models either rely on fully-observed training data or are suitable only for the MCAR data. Another recent approach, Distribution Matching (DM) \cite{muzellec_2020}, bypasses direct modeling of data distributions. A notable DM method is Transformed Distribution Matching \cite{zhao_2023_TDM}, which is suitable for real-world data with complex geometry. Non-parametric methods like k-nearest neighbors (k-NN) imputation \cite{troyanskaya_2001_knn}, which is a discriminative method, and MissForest \cite{stekhoven_2012}, which is an iterative and discriminative method, have shown effectiveness compared to sophisticated methods \cite{emmanuel_2021, jager_2021}, particularly in the MAR setting \cite{jarrett_2022}. Additionally, simple imputation approaches like mean substitution \cite{hawthorne_2005} provide basic alternatives. More details are provided in Appendix \ref{appendix:related-work}.




\section{Experiments}
\label{experiments}

We systematically assess CRILM's efficacy in addressing the research questions outlined in Section \ref{intro} through a series of experiments. Utilizing two types of LMs—decoder-only and encoder-decoder—we evaluate the performance of LMs fine-tuned with missingness-aware contextual datasets in downstream classification tasks post-imputation. Specifically, we investigate three types of missingness mechanisms: MCAR, MAR, and MNAR. For comparison with the baseline methods, we first impute the numeric datasets using existing methods (described further below). Then, the datasets are transformed into contextualized natural language datasets using the method described in Section \ref{CRILM}, which are used for fine-tuning the LMs.

\begin{figure*}[!htb]
    \begin{center}
        \subfigure[MCAR (Llama)]{\label{fig:Comparison-MCAR-Static-Llama}\includegraphics[width=0.5\textwidth]{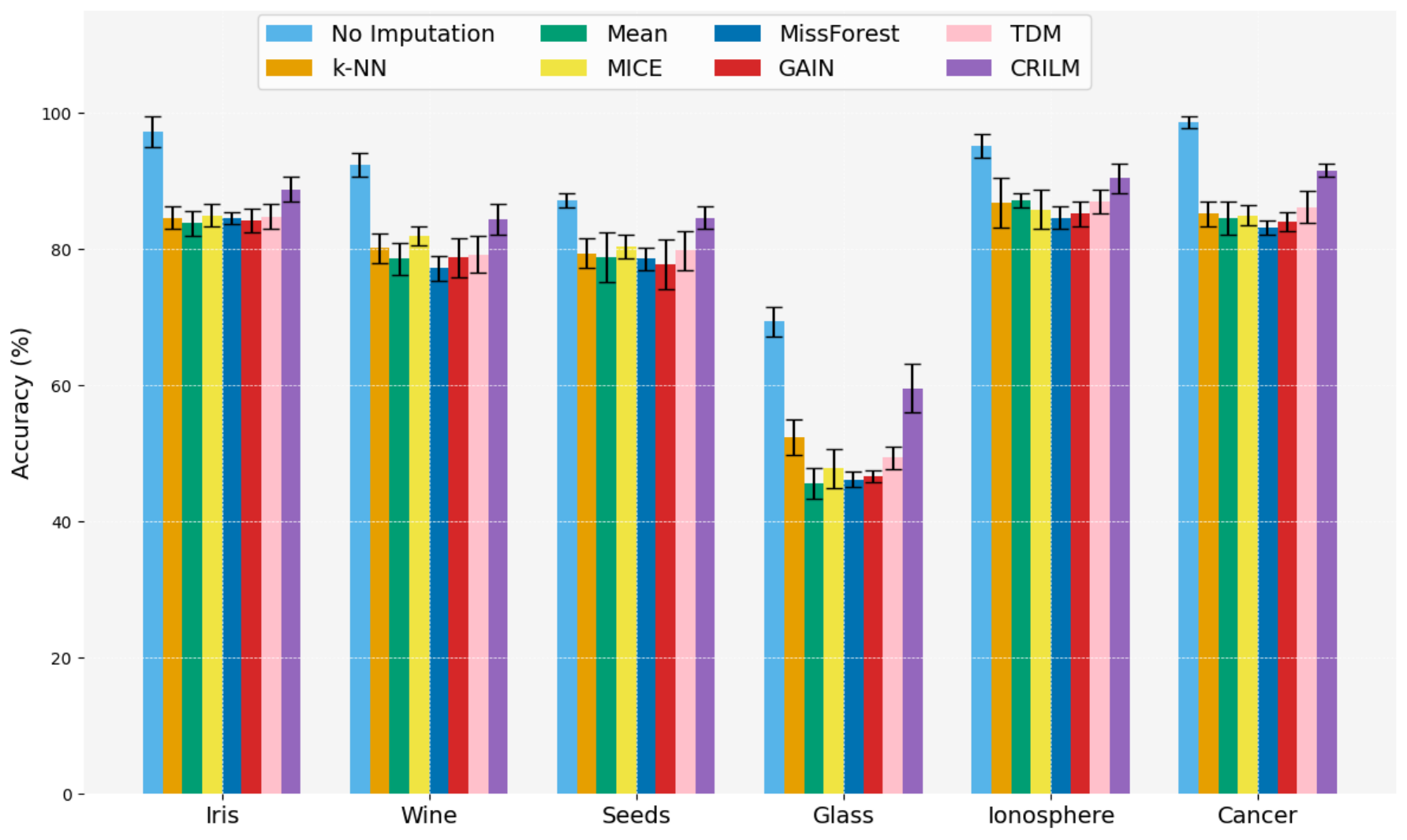}} \quad \hspace{-6mm}
        \subfigure[MCAR (FLAN-T5)]{\label{fig:Comparison-MCAR-Static-FLAN}\includegraphics[width=0.5\textwidth]{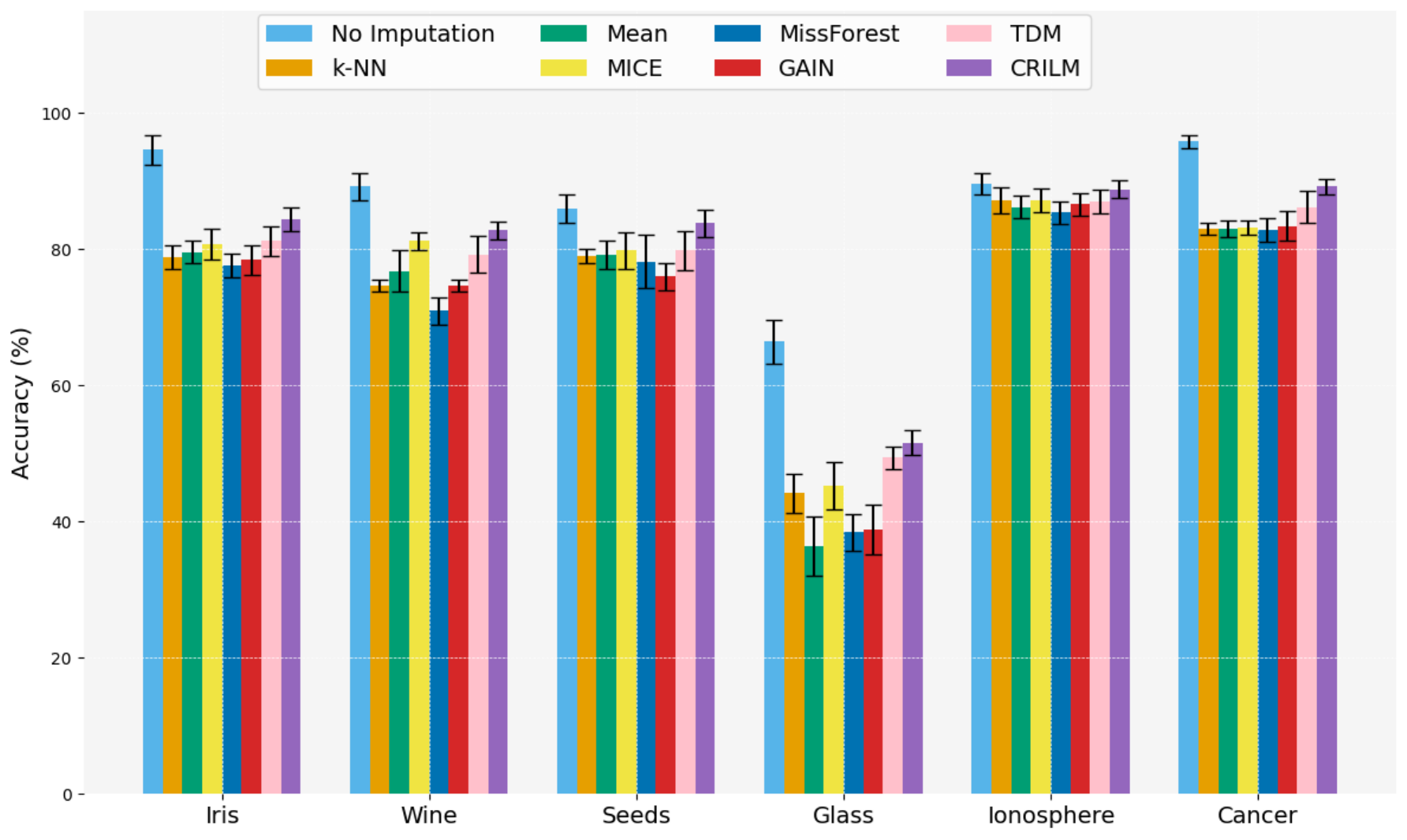}} \\ \vspace{-3mm}
        \subfigure[MAR (Llama)]{\label{fig:Comparison-MAR-Static-Llama}\includegraphics[width=0.5\textwidth]{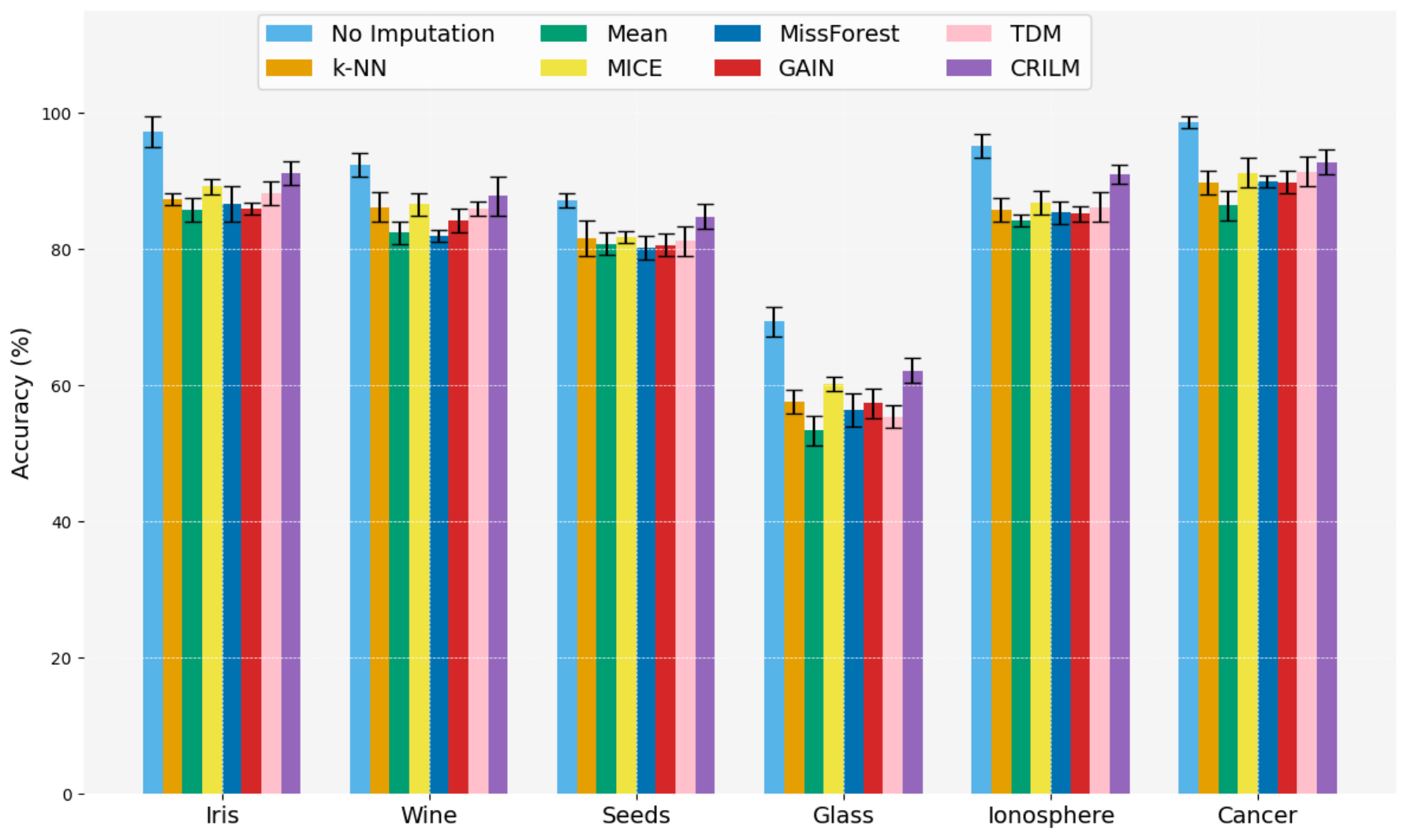}} \quad \hspace{-6mm}
        \subfigure[MAR (FLAN-T5)]{\label{fig:Comparison-MAR-Static-FLAN}\includegraphics[width=0.5\textwidth]{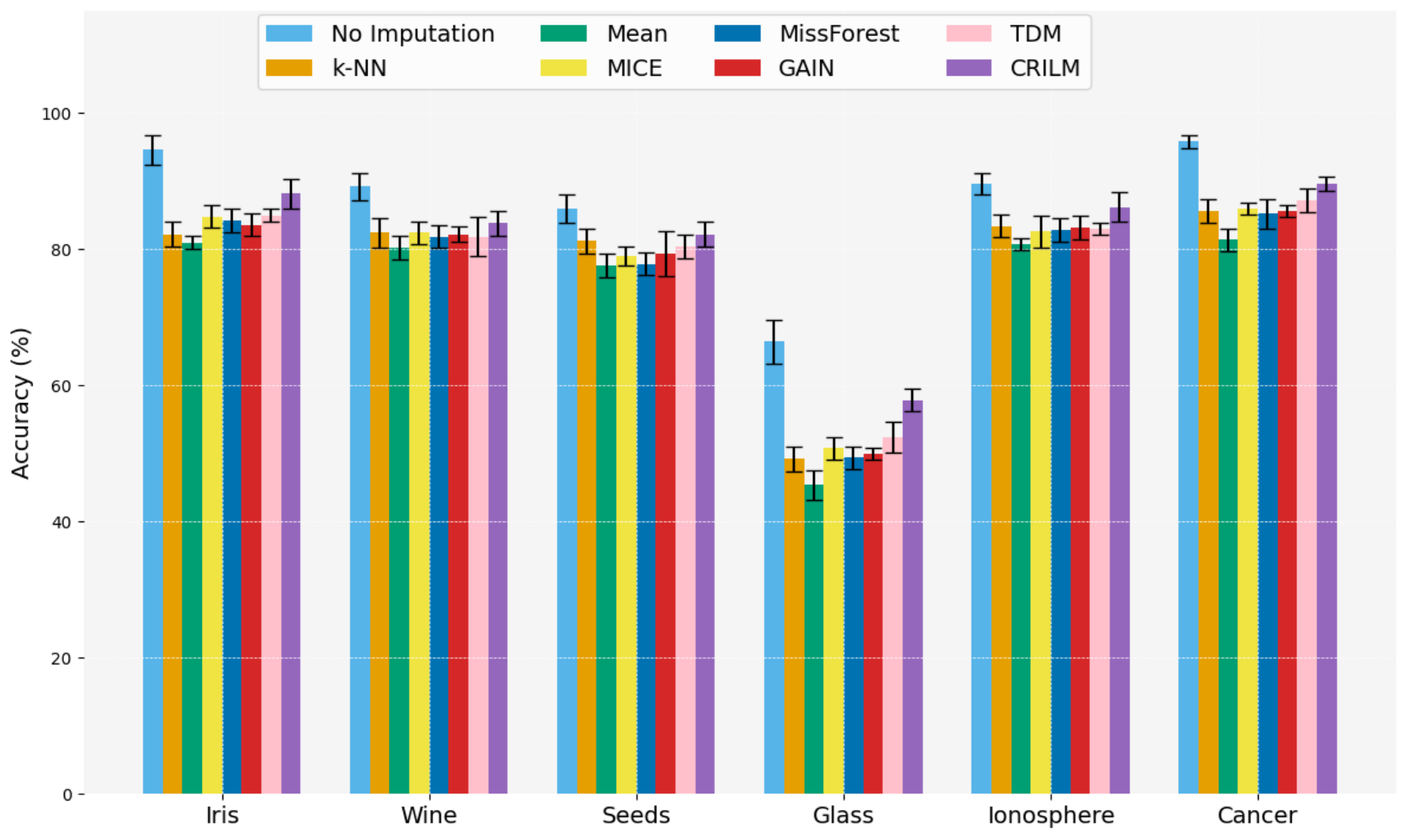}} \\ \vspace{-3mm}
        \subfigure[MNAR (Llama)]{\label{fig:Comparison-MNAR-Static-Llama}\includegraphics[width=0.5\textwidth]{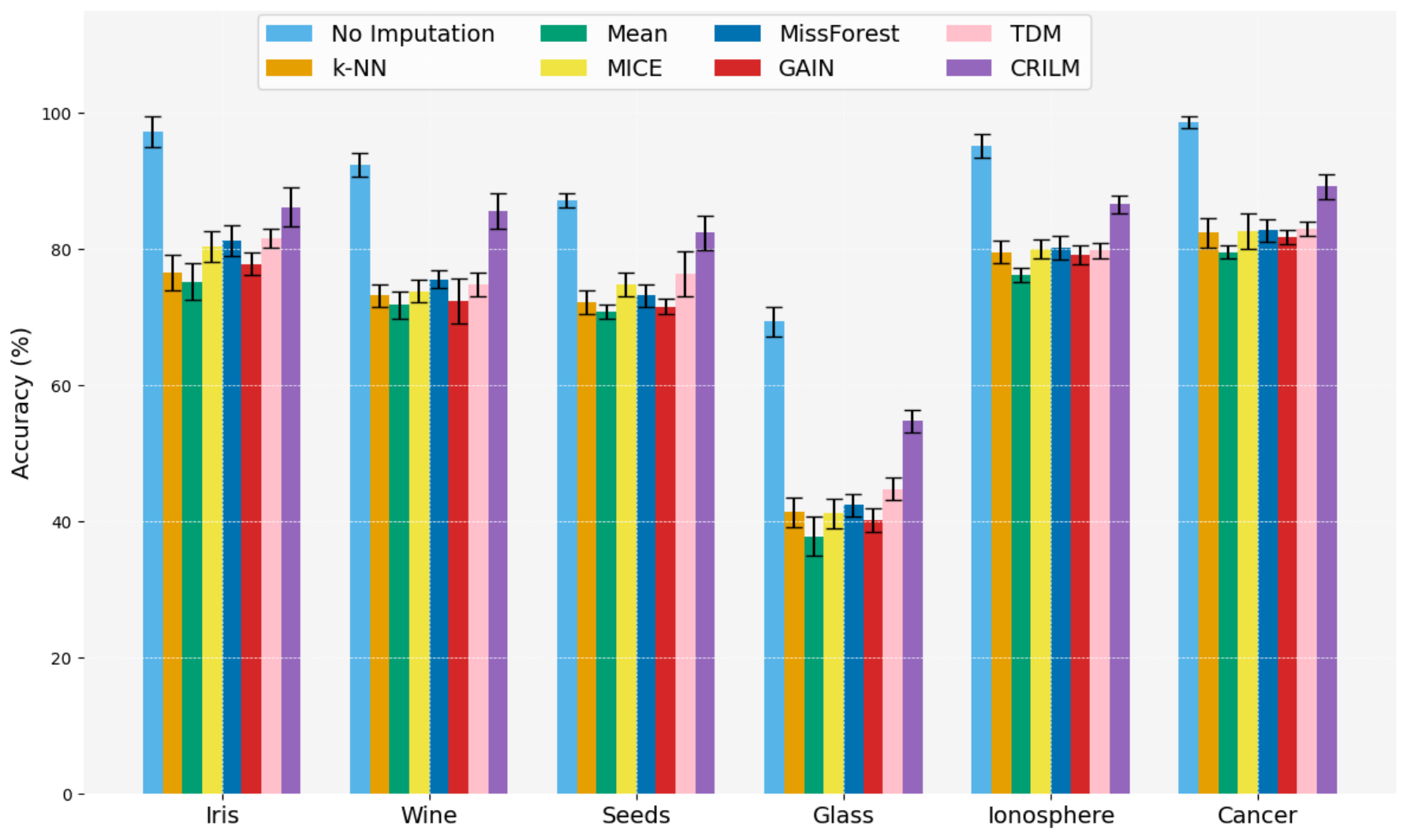}} \quad \hspace{-6mm}
        \subfigure[MNAR (FLAN-T5)]{\label{fig:Comparison-MNAR-Static-FLAN}\includegraphics[width=0.5\textwidth]{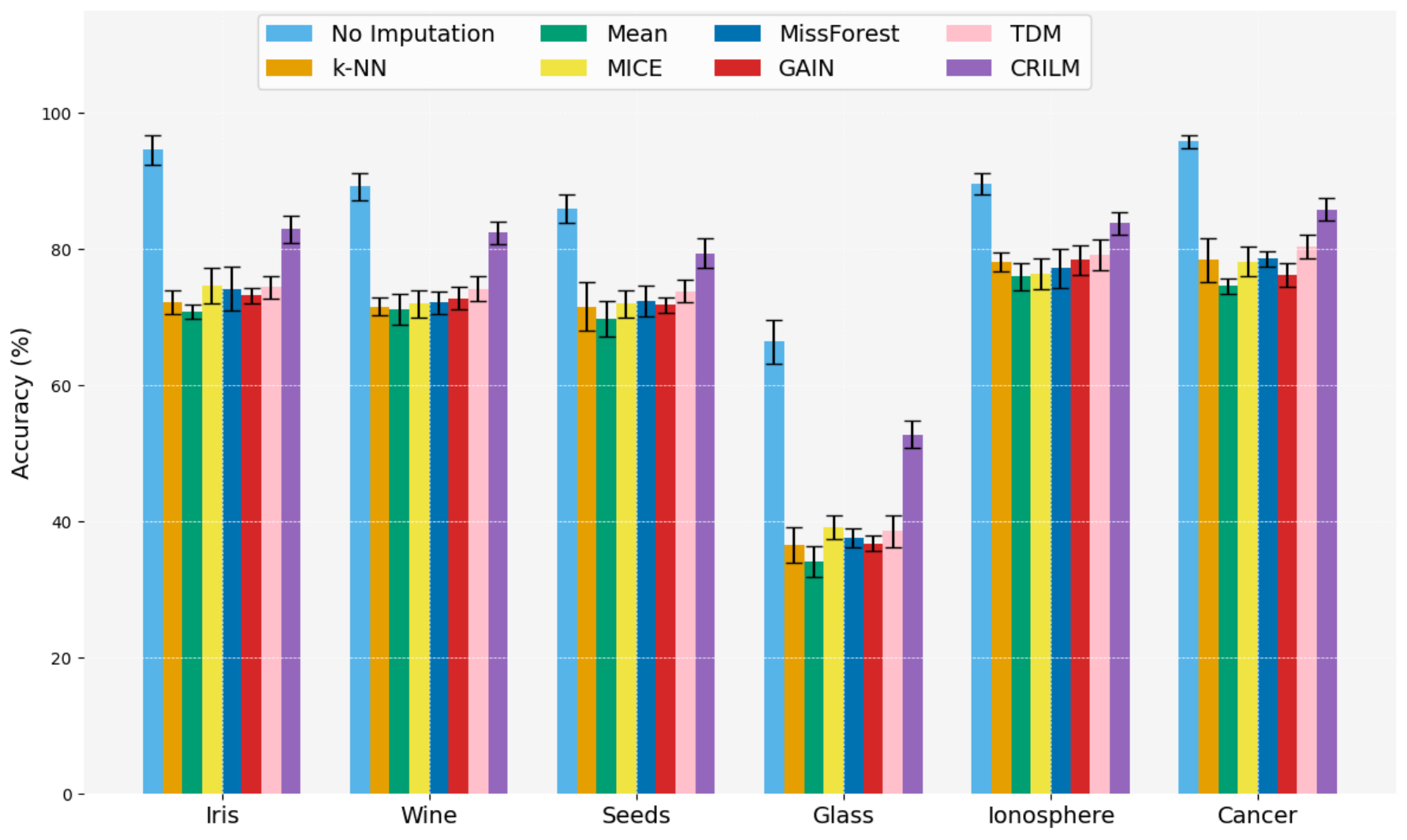}} \\ \vspace{-4mm}

         \caption{\textbf{[RQ1]}: Comparison of CRILM and baseline imputation methods across MCAR, MAR, and MNAR missingness patterns using Llama and FLAN-T5 models. Evaluation involves post-imputation LM-based downstream task performance, with CRILM fine-tuned on missingness-aware contextual datasets and baseline methods on contextual datasets. ``No Imputation'' cases show LM performance on complete datasets without missing values.}
        \label{fig:Comparison-CRILM-Static}
    \end{center}
    \vspace{-7mm}
\end{figure*}



\vspace{1.00mm}
\noindent \textbf{Datasets.}
We evaluate CRILM's performance using six real-life multivariate classification datasets from the UCI repository \cite{dua_2017_UCI}, which are selected based on their prior usage in existing numeric imputation-based studies \cite{muzellec_2020, yoon_2018_gain_PMLR, camino_2019, gondara_2018_MIDA, lu_2020, hallaji_2021_DLIN, nazabal_2018, zhao_2023_TDM}. This selection ensures a fair comparison with previous research efforts. Dataset statistics are provided in Appendix \ref{appendix:datasets-summary}.


\vspace{1.00mm}
\noindent \textbf{Baseline Imputation Methods.}
We compare CRILM against a diverse set of imputation approaches by focusing on the following six baseline methods: (1) Mean substitution \cite{hawthorne_2005} (simple imputation method), (2) k-NN \cite{troyanskaya_2001_knn, batista_2002} (non-parametric and discriminative method), (3) MissForest \cite{stekhoven_2012} (non-parametric, discriminative, and iterative method), (4) MICE \cite{vanBuuren_2006, vanBuuren_2011} (discriminative and distribution modeling iterative approach that treats each feature separately), (5) GAIN \cite{yoon_2018_gain_PMLR} (generative and distribution modeling iterative approach that treats features jointly), and (6) Transformed Distribution Matching (TDM) \cite{zhao_2023_TDM} (distribution matching method). 

\vspace{1.00mm}
\noindent \textbf{LMs for Downstream Tasks.}
We utilize two types of smaller pre-trained LMs for transfer learning: decoder-only Llama 2 \cite{touvron_2023_llama} and encoder-decoder FLAN-T5 \cite{chung_scaling_2022} with 7 billion (7B) and 770 million (770M) parameters, respectively.


\vspace{1.00mm}
\noindent \textbf{Experimental Settings.} The hyperparameter settings for the various imputation methods and the LMs used in our experiments are detailed below.

\vspace{1.00mm}
\textit{Hyperparameters for Baseline Imputation Methods.}
For GAIN, we adhere to the hyperparameters specified in the original publication, setting $\alpha$ to 100, the batch size to 128, the hint rate at 0.9, and the number of iterations to 1000 for optimal performance. MissForest and MICE are configured with their respective default parameters as provided in their PyPI implementations\footnote{https://pypi.org/}, i.e., MissForest: maxiter = 10, ntree = 100, and MICE: m = 5 for the number of multiple imputations. The PyPI MICE implementation utilizes random forests for efficiency. For k-NN, we determine the optimal values for $k$ for each dataset through hyperparameter tuning based on the downstream classification task. For a list of optimal $k$ values, refer to the Appendix \ref{appendix:optimal-k-values}. Regarding TDM, we use the original implementation with the reported settings \cite{zhao_2023_TDM}.

\vspace{1.00mm}
\textit{Pre-trained LMs for Transfer Lerning.} The Llama model is fine-tuned with the parameter-efficient QLoRA method \cite{dettmers_2023_qlora}. The settings are $r = 16$, $\alpha = 64$, $dropout = 0.1$ with the task type set to ``CAUSAL\_LM''. The learning rate is 2e-4, using the ``paged\_adamw\_32bit'' optimizer. The FLAN-T5 model \cite{chung_scaling_2022} is fine-tuned using an AdamW optimizer \cite{loshchilov_2018} with a learning rate set to 3e-4. 

Experiments are conducted with a batch size of 4 across 50 epochs, considering memory constraints during fine-tuning. Two Tesla A40 GPUs are used for distributed training, ensuring efficient processing, with each experiment completing in less than twenty minutes, except for the Breast Cancer dataset with more than 500 instances, which takes about an hour.  An estimated training time on a single GPU would require between 45 minutes to 2 hours to complete all experiments. For evaluation, 20\% of instances are randomly sampled from each dataset. Models are evaluated five times, and both the average performance and standard deviation are reported for comprehensive analysis.



\subsection{Results}

Figure \ref{fig:Comparison-CRILM-Static} displays experimental outcomes using two types of downstream LMs across six datasets, benchmarking CRILM against existing imputation methods. Performance metrics for LMs fine-tuned on complete datasets (without missing values, thus no imputation needed) are included for comparison. This approach highlights CRILM's effectiveness by providing a reference baseline, offering a clear view of its advantages over traditional imputation methods.


\vspace{1.00mm}
\noindent \textbf{[RQ1]}: \textit{To what extent does CRILM effectively perform in imputing missing values across distinct missingness mechanisms (MCAR, MAR, and MNAR), compared to existing methods, in terms of accuracy and robustness on varied datasets?}

\textbf{MCAR}: CRILM demonstrates superior accuracy in imputing missing values across all datasets compared to baseline imputation methods. Both the Llama and FLAN-T5 performed well, with Llama showing a slight advantage (1 to 8\% higher accuracy). CRILM's performance under the MCAR assumption, where missingness is independent of any data, suggests that it efficiently leverages contextual information for imputation. This efficacy is particularly evident in its ability to significantly close the gap toward the performance of fully complete datasets, showcasing its effectiveness.


\textbf{MAR}: CRILM's adaptability is further highlighted under MAR, where missingness depends on observed data. It outperforms other methods by a considerable margin, indicating its proficiency in utilizing available data points to predict missing values accurately. The Llama consistently exhibits superior performance, similar to the MCAR case (2 to 5\% higher accuracy).


\textbf{MNAR}: The MNAR scenario, characterized by missingness that depends on unobserved data, poses the most significant challenge. Here, CRILM's performance remains notably superior to traditional imputation methods. This robustness in the face of the most difficult missingness mechanism illustrates CRILM's potential to effectively mitigate biases introduced by MNAR missingness, utilizing the LMs' capacity to infer missing information from complex patterns. Similar to the previous cases, Llama exhibits better performance (2 to 4\% higher accuracy) 


\vspace{1.00mm}
To further demonstrate CRILM's superior performance over traditional baseline imputation methods, particularly in the \textbf{MNAR setting}, we assess its efficacy on three challenging datasets: Glass Identification, Seeds, and Wine. These datasets are selected due to the observed lower performance of LMs when utilizing fully complete versions (refer to Figure \ref{fig:Comparison-CRILM-Static}), highlighting their complexity and serving as a rigorous evaluation benchmark for CRILM. According to the results (see Table  \ref{tab:comparison_three_datasets}), CRILM consistently outperforms the best baseline methods. The performance gains are 10.0\%, 6.0\%, and 10.0\% for Glass Identification, Seeds, and Wine, respectively, using Llama, and 13.6\%, 5.6\%, and 8.2\% using FLAN-T5. This significant improvement underscores CRILM's effectiveness in addressing the intricacies of MNAR missingness, confirming its position as a robust tool for managing various missing data scenarios. Additional analysis details on MCAR and MAR are provided in Appendix \ref{appendix:performance}.


\vspace{-2.00mm}
\begin{table}[!htb]
    
   \label{tab:comparison_three_datasets}
   \vspace{-3.00mm}
    {\small
     \begin{tabular}{p{1.19cm} p{0.65cm} p{2.45cm} p{0.7cm} p{0.65cm}}
        \toprule
        \textbf{LM} & \textbf{Data} & \textbf{Best Baseline} & \textbf{CRILM} & \textbf{Gain} \\
        \midrule
        & Glass & 44.80\% (TDM) &  54.80\% & +10.0\% \\
        Llama & Seeds & 76.40\% (TDM) & 82.40\% & +6.0\% \\
        & Wine & 75.60\% (MissForest) & 85.60\% & +10.0\% \\
        \midrule\midrule
        & Glass & 39.20\% (MICE) & 52.80\% & +13.6\% \\
        FLAN-T5 & Seeds & 73.80\% (TDM) & 79.40\% & +5.6\% \\
        & Wine & 74.20\% (TDM) & 82.40\% & +8.2\% \\
        \bottomrule
    \end{tabular}}
    \centering
    \caption{Comparison of CRILM accuracy with leading imputation methods on MNAR missingness across three datasets. }
\end{table}

\begin{figure*}[!htb]
    \begin{center}
        \subfigure[MCAR (Llama)]{\label{fig:Comparison-Descriptors-MCAR-Static-Llama}\includegraphics[width=0.5\textwidth]{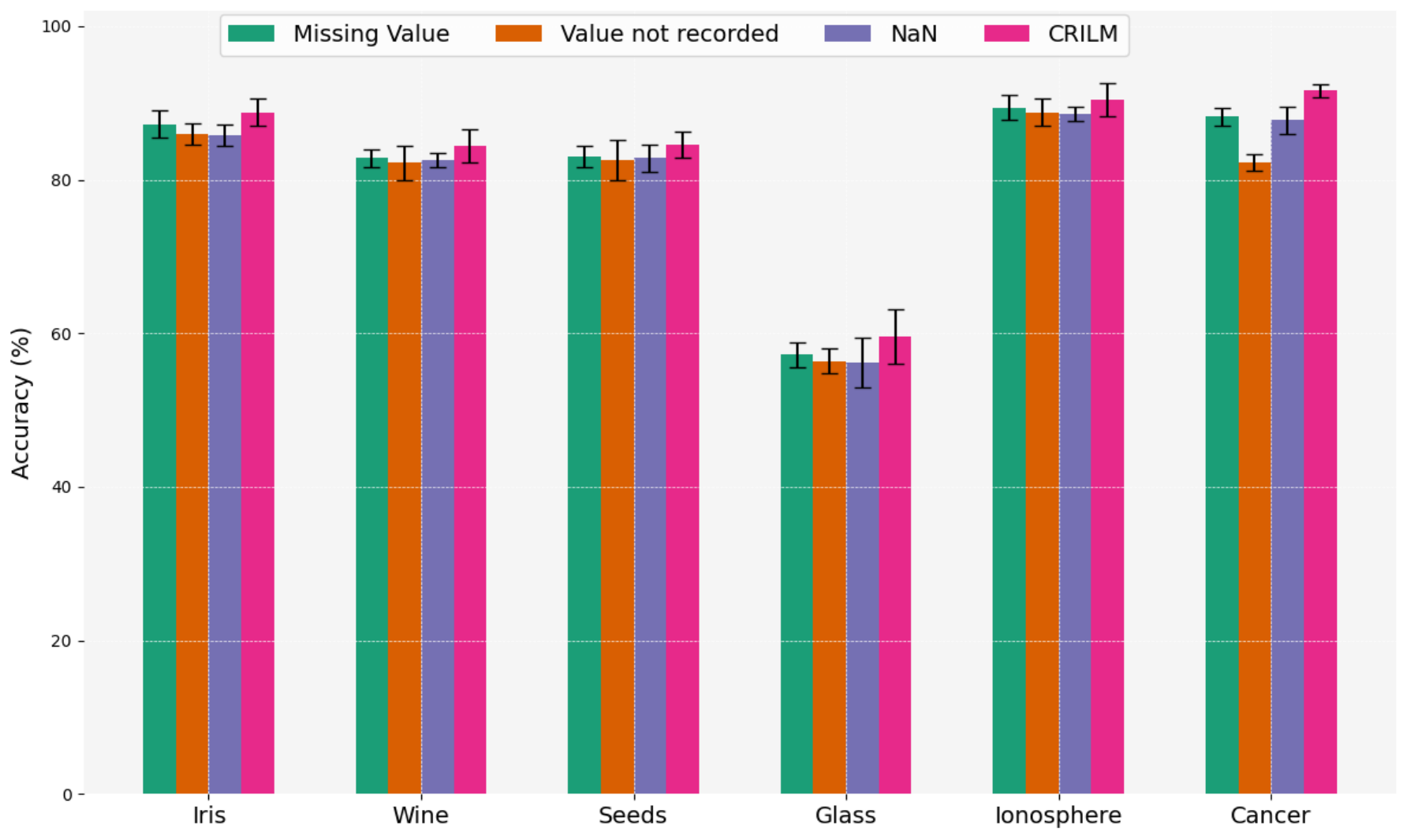}} \quad  \hspace{-6.00mm}
        \subfigure[MCAR (FLAN-T5)]{\label{fig:Comparison-Descriptors-MCAR-Static-FLAN}\includegraphics[width=0.5\textwidth]{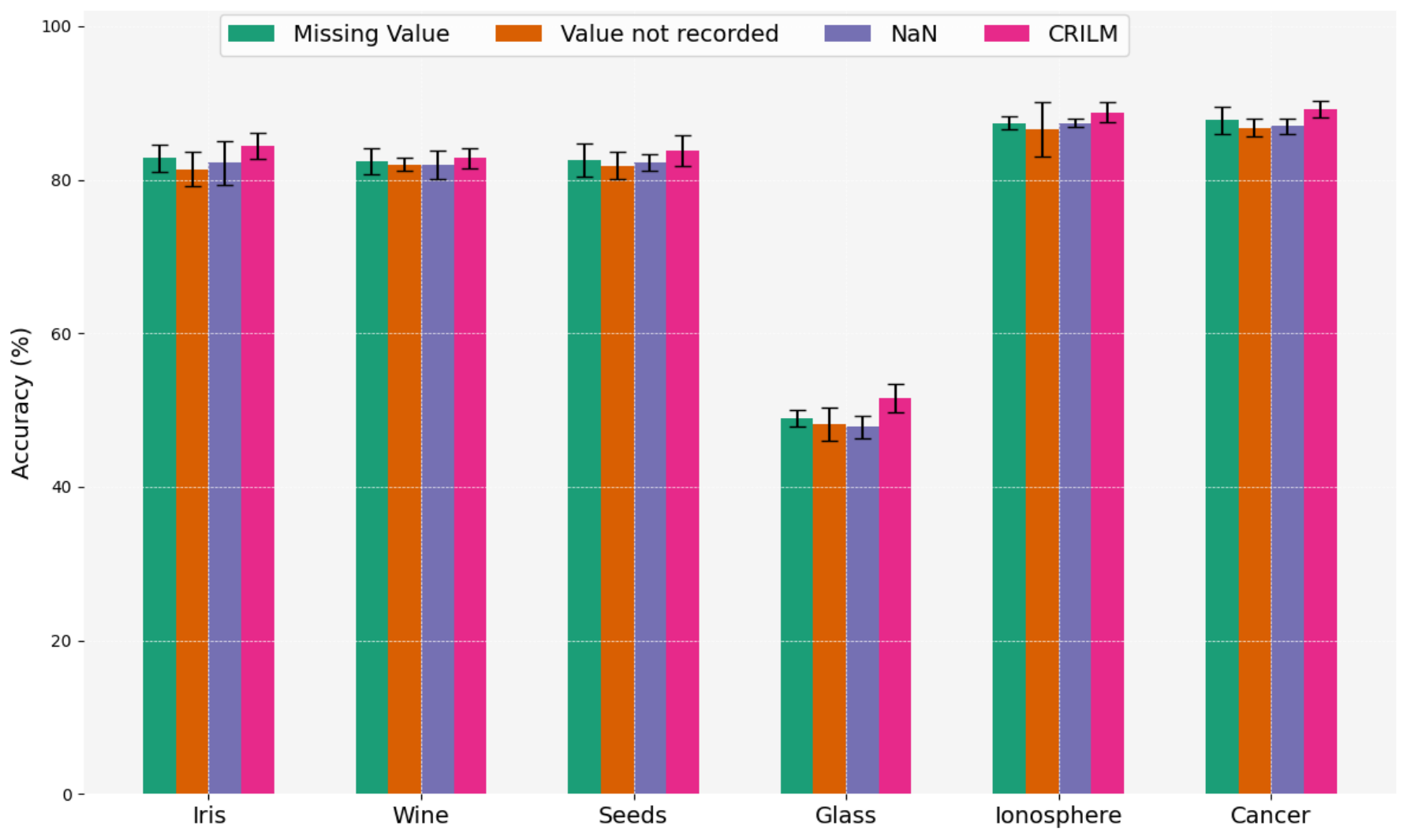}} \\  \vspace{-4.00mm}
    \caption{\textbf{[RQ2]}: Impact of feature-specific vs. generic (``NaN'', ``Missing value'', and ``Value not recorded'') missingness descriptors on LM Performance in MCAR scenario.}
    
    \label{fig:Comparison-Descriptors-MCAR-Static}

    \end{center}
    \vspace{-6.00mm}
\end{figure*}

\vspace{1.00mm}
\noindent \textbf{Discussion on RQ1.} CRILM's consistent superiority across diverse missingness patterns and datasets confirms its effectiveness, \textbf{addressing RQ1}. This underscores the advantages of integrating contextualized natural language models into imputation, particularly in challenging MNAR scenarios where traditional numeric-estimation methods may introduce biases. The robust performance of CRILM across MCAR, MAR, and MNAR missingness mechanisms highlights its broad applicability, distinguishing it from conventional methods. This generalizability can be attributed to CRILM's missingness-aware data contextualization approach, which effectively taps into the prior knowledge of the pre-trained LMs to implicitly handle missing cases in the data. Notably, Llama (7B) performs slightly better than FLAN-T5 (770M), likely due to its larger model size, which enhances its ability to capture and utilize complex patterns in the data. Furthermore, minimal performance variation across iterations underscores CRILM's stability and reliability, crucial for real-world applications. Its ability to maintain a consistently low error margin highlights its potential as a reliable solution for data imputation.

\vspace{1.00mm}
\noindent \textbf{[RQ2]}: \textit{How do feature-specific versus generic missingness descriptors impact the performance of LM-based downstream tasks?} Initially, we utilize contextually relevant, feature-specific descriptors for missing values, leading to unique phrases for different features within a dataset. To address RQ2, we aim to determine whether using a uniform, yet contextually relevant, descriptor for all features would offer comparable benefits. To this end, we experiment with three consistent descriptors: ``NaN'', ``Missing value'', and ``Value not recorded''. These experiments, focusing on the MCAR scenario, sought to ascertain whether it is more beneficial to use contextually nuanced descriptors or whether a generic descriptor is adequate to harness LMs' general knowledge for managing missing values in datasets.

The experimental findings (Figure \ref{fig:Comparison-Descriptors-MCAR-Static}) illuminate the influence of missing data phrasing on the effectiveness of LMs in addressing such situations. The results reveal a distinct pattern across both types of LMs: generic descriptors, such as ``NaN'', consistently perform worse than context-specific descriptors designed for each feature and dataset. Among the three fixed descriptors tested, there are some variations in performance. Both ``NaN'' and ``Missing value'' outperformed ``Value not recorded'', with ``Missing value'' achieving the best results in most cases among the static descriptors.



\vspace{2.00mm}
\noindent \textbf{Discussion on RQ2.}  The findings on RQ2 highlight the importance of context in LMs' handling of missing data. The superior performance of feature-specific descriptors shows that LMs better manage missing data when it is described in a way that accurately reflects the context of the missing information. For example, a descriptor like \textit{``Malic acid quantity missing for this wine sample''} allows an LM to interpret and address the missing data point more effectively than a generic descriptor like \textit{``The level of malic acid in the wine is NaN''}. This preference for context-specific descriptors stems from LMs' extensive linguistic capability. When missing data aligns with the specific context of a feature, an LM can better utilize its knowledge to handle the missing values. However, effectiveness drops when generic labels are used, as they provide minimal contextual information for the LM to draw upon.



\vspace{2.00mm}
\noindent \textbf{Cost-Effective Implementation of CRILM.} Our method provides an economically viable solution by utilizing publicly available LLMs for zero-shot inference and smaller LMs for downstream tasks, allowing for efficient fine-tuning even in resource-constrained settings. This feasibility is demonstrated through experiments employing accessible resources like ChatGPT-3.5 for inference and single GPU fine-tuning, ensuring experiments are completed within an hour on average, thereby highlighting its cost-effectiveness.

\section{Conclusion}
\label{conclusion}

CRILM demonstrates robust handling of missing data across MCAR, MAR, and notably MNAR mechanisms, consistently outperforming traditional methods. Our experiments highlight CRILM's remarkable effectiveness in MNAR scenarios, achieving up to a 10\% performance margin over baseline methods, underscoring its efficacy in the most challenging missingness setting. By leveraging contextualized LMs, CRILM offers a novel imputation method alongside numeric-estimation approaches, particularly beneficial in mitigating biases and enhancing reliability in MNAR case. Its cost-effective implementation, using publicly available LLMs for inference and smaller LMs for downstream tasks, enhances practicality in resource-constrained settings.

Future work will explore extending CRILM to diverse data types such as time-series, images, and unstructured text.

\section{Limitations}
\label{limitations}

Despite the notable advancements presented by CRILM in addressing missing data within tabular datasets, this work has several limitations. Firstly, CRILM's efficacy depends heavily on the quality and diversity of the training data used to develop the underlying LLMs. In scenarios where LLMs lack exposure to data similar to the specific domain or context of missing information, their ability to generate accurate imputations may be compromised. Additionally, the approach assumes that the descriptive context provided for missing values sufficiently informs the LLM, which may not always be the case. 
Furthermore, processing large datasets with CRILM, even though we utilize smaller LMs for fine-tuning with contextualized missingness-aware data, may pose scalability challenges, as the fine-tuning process could increase in duration. Moreover, while CRILM performs well across various missingness mechanisms, its application in highly specialized domains where expert knowledge heavily influences data interpretation requires further exploration. Lastly, it is important to note that our evaluation focused on classifying downstream tasks, leaving its efficacy in other task types for future investigation.

\section*{Acknowledgments}
This research was supported by a grant from the U.S. National Science Foundation (NSF DUE 2142558).

\bibliography{references}

\appendix

\section{Appendix}
\label{appendix}


In this section, we begin with a comprehensive discussion of the related work. Following this, we conduct a comparative analysis of CRILM's effectiveness on a selected set of challenging downstream tasks. Next, we provide a summary of the datasets, along with a list of feature-specific contextually relevant missing-value descriptors for three selected datasets. Lastly, we present the optimal values of $k$ obtained through hyperparameter tuning for k-NN imputation across three missingness patterns—MCAR, MAR, and MNAR—using the Llama and FLAN-T5 models on the six datasets.

\subsection{Related Work}
\label{appendix:related-work}

The challenge posed by missing data in tabular datasets has led to the development of numerous imputation methods, broadly classified into two categories: those modeling feature distribution and those that do not model distributions. The latter category includes methods such as distribution matching and traditional non-parametric methods. In the former category, where the focus is on modeling feature distribution, methods aim to model the distribution of missing values while maximizing the observed likelihood \cite{muzellec_2020}. Within this line of approach, two distinct types of imputation methods exist \cite{zhao_2023_TDM}: methods that treat features separately and those that treat them jointly.

For methods treating features separately, an \textbf{iterative approach} is employed, specifying a univariate model for each feature based on all others. A prominent example is the Multivariate Imputation by Chained Equations (MICE) method \cite{vanBuuren_2006, vanBuuren_2011}, which adopts a \textit{discriminative approach} for imputation. MICE sequentially imputes missing values for each variable based on the others, cycling through the variables iteratively until predictions stabilize. MICE is particularly effective for handling MCAR and MAR data \cite{jarrett_2022}. Other notable iterative methods include \cite{heckerman_2000, raghunathan_2001, gelman_2004, liu_2014, zhu_2015}. Given the potential variation in the conditional distribution of each feature, these methods necessitate the specification of separate models for each feature. This approach may prove ineffective, especially in cases where the nature of the missing values remains uncertain.

On the other hand, methods that treat features collectively aim to learn a joint distribution of all features, either explicitly or implicitly. A classical approach for explicit joint modeling assumes a Gaussian distribution for the data, with parameters estimated using EM algorithms \cite{dempster_1977_EM}. Recent developments have seen the utilization of deep learning-based \textit{generative methods} such as Denoising Autoencoders (DAE) \cite{vincent_2008} and Generative Adversarial Nets (GAN) \cite{goodfellow_2014_generative}. Generative methods can be categorized into implicit and explicit modeling. Implicit models include imputers trained as generators in GAN-based frameworks \cite{yoon_2018_gain_PMLR, li_2019_MisGAN, yoon_2020_GAMIN, dai_2021, fang_2022_FragmGAN}. However, these models produce imputations that are only valid for the MCAR data \cite{yoon_2018_gain_PMLR, li_2019_MisGAN, yoon_2020_GAMIN}. A notable GAN-based method is GAIN \cite{yoon_2018_gain_PMLR}, specifically designed for imputing missing data without the need for complete datasets. In GAIN, the generator outputs the imputations, while the discriminator classifies the imputations on an element-wise basis. However, GAIN can be quite difficult to implement in practice \cite{muzellec_2020}. Moreover, it often falls short compared to more traditional machine learning methods such as the non-parametric k-nearest neighbors (k-NN) in terms of performance~\cite{jager_2021}. Explicit generative models refer to deep latent-variable models trained to approximate joint densities using variational bounds. Most of these models either rely on fully-observed training data \cite{gondara_2018_MIDA, rezende_2014, mattei_2018} or are suitable only for the MCAR data \cite{nazabal_2020, ivanov_2018, richardson_2020}. MIWAE \cite{mattei_2019_MIWAE} is an exception in this category that adapts the importance-weighted autoencoders \cite{burda_2015} objective to approximate maximum likelihood in MAR settings. However, its accuracy depends on the assumption of infinite computational resources. Additionally, with the exception of methods that use separate decoders for each feature \cite{nazabal_2020}, generative methods generally do not support column-specific modeling. Other approaches in the category of methods that learn a joint distribution include those based on matrix completion \cite{mazumder_2010, hastie_2015}, graph neural networks \cite{you_2020, vinas_2021, chen_2022_GEDI, huang_2022_graph, moralesAlvarez_2022, gao_2023_graph}, normalizing flows \cite{richardson_2020_MCFLOW, ma_2021_EMFlow, wang_2022_flow_imputation}, and Gaussian processes \cite{dai_2022_multiple_imputation}.

Distribution Matching (DM) methods represent a recent alternative approach that bypasses the need for modeling data distributions directly \cite{muzellec_2020, zhao_2023_TDM}. The core idea behind DM is that any two batches of data (including those with missing values) originate from the same underlying data distribution. Therefore, an effective method should impute the missing values to ensure that the empirical distributions of the two batches are closely matched. In \cite{muzellec_2020}, the authors achieve DM by minimizing the optimal transport (OT) distance, with the cost function being the quadratic distance in the data space between samples. Another notable method, suitable for real-world data with complex geometry, is Transformed Distribution Matching (TDM) \cite{zhao_2023_TDM}. TDM performs OT-based imputation in a transformed space, where the distances between transformed samples better reflect their underlying similarities and dissimilarities, respecting the data's inherent geometry.

Non-parametric methods like k-NN imputation \cite{troyanskaya_2001_knn, batista_2002} and random forest imputation, such as MissForest \cite{stekhoven_2012}, have demonstrated effectiveness in comparison to other sophisticated imputation methods \cite{emmanuel_2021, jager_2021}. The k-NN method employs a discriminative algorithm that utilizes the similarity between instances, typically measured by Euclidean distance, to impute missing values, offering flexibility in handling both continuous and categorical data. Conversely, MissForest is an \textit{iterative method} harnessing the power of random forests, excelling in datasets with complex interactions and non-linear relationships, often surpassing other methods in terms of accuracy and robustness. MissForest is particularly adept in the MAR setting \cite{jarrett_2022}.

Finally, simple imputation approaches like mean substitution \cite{hawthorne_2005} and hot deck imputation \cite{marker_2002} provide basic alternatives.

\subsection{Comparative Analysis of CRILM's Effectiveness}
\label{appendix:performance}

\begin{table}[htb!]

\label{tab:comparison_three_datasets_llama}
{\small
\begin{tabular}{p{1.0cm} p{2.7cm} p{1.0cm} p{0.8cm}}
\toprule
\textbf{Dataset} & \textbf{Best Baseline} & \textbf{CRILM} & \textbf{Gain} \\
\midrule
\multicolumn{4}{c}{\textbf{Glass Identification}} \\
\midrule
MCAR & 52.40\% \textbf{(k-NN)} & 59.60\% & 7.2\% \\
MAR & 60.20\% \textbf{(MICE)} & 62.20\% & 2.0\% \\
MNAR & 44.80\% \textbf{(TDM)} & 54.80\% & 10.0\% \\
\midrule
\multicolumn{4}{c}{\textbf{Seeds}} \\
\midrule
MCAR & 80.40\% \textbf{(MICE)} & 84.60\% & 4.2\% \\
MAR & 81.80\% \textbf{(MICE)} & 84.80\% & 3.0\% \\
MNAR & 76.40\% \textbf{(TDM)} & 82.40\% & 6.0\% \\
\midrule
\multicolumn{4}{c}{\textbf{Wine Quality}} \\
\midrule
MCAR & 82.00\% \textbf{(MICE)} & 84.40\% & 2.4\% \\
MAR & 86.60\% \textbf{(MICE)} & 87.80\% & 1.2\% \\
MNAR & 75.60\% \textbf{(MissForest)} & 85.60\% & 10.0\% \\
\bottomrule
\end{tabular}
}
\centering
\caption{Performance Comparison of CRILM accuracy with leading imputation methods using Llama across three challenging datasets. Best performing baseline methods are in \textbf{bold}.}
\end{table}

To demonstrate the superior performance of CRILM over traditional baseline imputation methods, we investigate its performance on \textbf{three particularly challenging datasets}: Glass Identification, Seeds, and Wine. These datasets were chosen due to the comparatively lower performance exhibited by the LMs when using fully complete versions of the datasets (i.e., no missing values), underscoring their complexity and providing a rigorous testing ground for evaluating CRILM's effectiveness.



\subsubsection{Llama}

Table \ref{tab:comparison_three_datasets_llama} presents a detailed comparative analysis based on Llama. In the MCAR setting, CRILM demonstrates substantial superiority over the best baseline method (k-NN, achieving 52.40\% accuracy) with a performance gain of 7.2\%. This underscores CRILM's robustness in effectively handling missing data within complex datasets. The challenge intensifies with the Seeds dataset, where CRILM surpasses the top-performing baseline method (MICE) by 4.2\% under the MCAR setting. Similar trends are observed in the Wine dataset, where CRILM outperforms the best baseline performance under MCAR by 2.4\%.

Under MAR conditions, the performance gaps between CRILM and the best-performing baseline methods are relatively modest—2\%, 3\%, and 1.2\% for Glass Identification, Seeds, and Wine, respectively. This suggests that while the predictability of missingness from observed data in MAR scenarios provides some advantage to traditional imputation methods, CRILM still maintains a performance edge.

The MNAR scenario, characterized by the most complex pattern of missingness, highlights CRILM's distinct advantage. Across all three datasets, CRILM not only outperforms the best baseline methods but does so with remarkable performance gains of 10.0\%, 6.0\%, and 10\% for Glass Identification, Seeds, and Wine, respectively. This substantial improvement underscores CRILM's effectiveness in navigating the intricacies of MNAR missingness, further establishing its status as a robust tool for handling various missing data scenarios.

\subsubsection{FLAN-T5}

Table \ref{tab:comparison_three_datasets_flan} provides a FLAN-T5-based comparative analysis of CRILM against leading imputation methods across the three challenging datasets, echoing similar trends observed with the Llama model. In the Glass Identification dataset, FLAN-T5 exhibits significant improvements with CRILM. Under the MCAR setting, CRILM surpasses the best baseline method (TDM, achieving 45.60\% accuracy) by 6.0\%, highlighting its robust capability to handle missing data effectively, particularly where traditional methods struggle. The Seeds dataset presents a competitive landscape, where CRILM outperforms the top-performing baseline (MICE) by 4.0\% under MCAR conditions. Similarly, in the Wine Quality dataset under MCAR conditions, CRILM achieves a 1.2\% performance gain over MICE, reinforcing its reliability.

In the MAR scenario for Glass Identification, CRILM shows a pronounced advantage over the best baseline method (TDM, achieving 52.40\%), with a notable gain of 5.4\%. This underscores CRILM's efficacy in scenarios where missingness can be predicted from observed data, showcasing its versatility across different missing data patterns. However, in the challenging Seeds dataset, the performance gap narrows, with CRILM outperforming k-NN by 1.0\%, indicating its continued edge despite the predictability leveraged by traditional methods. The Wine Quality dataset reflects a similar trend, where CRILM achieves a 1.4\% performance gain over k-NN.

In the MNAR condition, known for its complexity, CRILM demonstrates a significant advantage. In the Glass Identification dataset, CRILM outperforms MICE by an impressive 13.6\%. This substantial improvement is mirrored in the Seeds and Wine Quality datasets, where CRILM achieves gains of 5.6\% and 8.2\% over TDM, respectively. These results underscore CRILM's exceptional capability in handling the intricate challenges posed by MNAR missingness, firmly establishing it as a powerful tool for addressing diverse imputation challenges.

\begin{table}[htb!]

\label{tab:comparison_three_datasets_flan}
{\small
\begin{tabular}{p{1.1cm} p{2.4cm} p{1.2cm} p{1.0cm}}
\toprule
\textbf{Dataset} & \textbf{Best Baseline} & \textbf{CRILM} & \textbf{Gain} \\
\midrule
\multicolumn{4}{c}{\textbf{Glass Identification}} \\
\midrule
MCAR & 45.60\% \textbf{(TDM)} & 51.60\% & 6.0\% \\
MAR & 52.40\% \textbf{(TDM)} & 57.80\% & 5.4\% \\
MNAR & 39.20\% \textbf{(MICE)} & 52.80\% & 13.6\% \\
\midrule
\multicolumn{4}{c}{\textbf{Seeds}} \\
\midrule
MCAR & 79.80\% \textbf{(MICE)} & 83.80\% & 4.0\% \\
MAR & 81.20\% \textbf{(k-NN)} & 82.20\% & 1.0\% \\
MNAR & 73.80\% \textbf{(TDM)} & 79.40\% & 5.6\% \\
\midrule
\multicolumn{4}{c}{\textbf{Wine Quality}} \\
\midrule
MCAR & 81.20\% \textbf{(MICE)} & 82.40\% & 1.2\% \\
MAR & 82.40\% \textbf{(k-NN)} & 83.80\% & 1.4\% \\
MNAR & 74.20\% \textbf{(TDM)} & 82.40\% & 8.2\% \\
\bottomrule
\end{tabular}
}
\centering
\caption{Performance Comparison of CRILM accuracy with leading imputation methods using FLAN-T5 across three challenging datasets. Best performing baseline methods are in \textbf{bold}.}
\end{table}

\subsection{Dataset Summary}
\label{appendix:datasets-summary}

Table \ref{tab:description_datasets} provides a summary of the six UCI datasets.

\begin{table*}[htb!]
\centering
\caption{Description of the datasets. N=size of the dataset and d=number of features.}
\label{tab:description_datasets}
\centering
\resizebox{\textwidth}{!}{%
\begin{tabular}{@{}lp{1cm}p{1cm}p{10cm}@{}} 
\toprule
Dataset & N & d & Description \\ \midrule
Iris & 150 & 4 & The dataset contains 3 classes of 50 instances each, referring to types of iris plants. \\
Wine & 178 & 13 & Results of a chemical analysis of wines grown in Italy, with three types represented. \\
Seeds & 210 & 7 & Properties of three varieties of wheat: Kama, Rosa, and Canadian. \\
Glass Identification & 214 & 9 & Classification of types of glass for criminological investigation. \\
Ionosphere & 351 & 34 & Phased array of 16 high-frequency antennas, targeting free electrons in the ionosphere. \\
Breast Cancer Wisconsin & 569 & 30 & Binary classification from digitized images of a fine needle aspirate of breast masses. \\ \bottomrule
\end{tabular}}
\end{table*}

\subsection{Missing-value Descriptors}
\label{appendix:descriptors}

Table \ref{tab:descriptors} reports the list of feature-specific contextually relevant missing-value descriptors for three selected datasets.

\begin{table*}[htb!]
\centering
\caption{Feature-specific contextually relevant descriptors for three selected datasets.}
\label{tab:descriptors}
\resizebox{\textwidth}{!}{%
\begin{tabular}{l|l|l}
\hline
Dataset & Features containing Missing   values                                                                                                                                                                                                                                                          & Descriptors of missing values                                                                                                                                                                                                                                                                                                                                                                                                                                                                                                                                                                                                                                                                                                                                                                                                                  \\ \hline
Iris    & \begin{tabular}[c]{@{}l@{}}1. Sepal Length\\ 2. Sepal Width\\ 3. Petal Length\\ 4. Petal Width\end{tabular}                                                                                                                                                                                   & \begin{tabular}[c]{@{}l@{}}1. Sepal Length: Unavailable \\ 2. Sepal Width: Unavailable \\ 3. Petal Length: Unavailable \\ 4. Petal Width: Unavailable\end{tabular}                                                                                                                                                                                                                                                                                                                                                                                                                                                                                                                                                                                                                                                                             \\ \hline
Wine    & \begin{tabular}[c]{@{}l@{}}1. Alcohol \\ 2. Malic acid \\ 3.Ash \\ 4. Alcalinity of ash \\ 5. Magnesium \\ 6. Total phenols \\ 7. Flavanoids \\ 8. Nonflavanoi phenols \\ 9. Proanthocyanins \\ 10. Color Intensity \\ 11. Hue \\ 12.OD280/OD315 of diluted wines \\ 13. Proline\end{tabular} & \begin{tabular}[c]{@{}l@{}}1. Alcohol content not provided for this wine sample. \\ 2. Malic acid quantity missing for this wine sample. \\ 3. Ash content data not recorded for this wine sample. \\ 4. Alcalinity of ash information unavailable for this wine sample. \\ 5. Magnesium level not specified for this wine sample. \\ 6. Total phenols data missing for this wine sample. \\ 7. Flavanoids content not available for this wine sample. \\ 8. Nonflavanoid phenols quantity not provided for this wine sample. \\ 9. Proanthocyanins data missing for this wine sample. \\ 10. Color intensity information not recorded for this wine sample. \\ 11. Hue value not specified for this wine sample. \\ 12. OD280/OD315 data missing for this wine sample. \\ 13. Proline content not available for this wine sample\end{tabular} \\ \hline
Seeds   & \begin{tabular}[c]{@{}l@{}}1. Area \\ 2. Perimeter \\ 3. Compactness \\ 4. Length of kernel \\ 5. Width of kernel \\ 6. Asymmetry coefficient \\ 7. Length of kernel groove\end{tabular}                                                                                                      & \begin{tabular}[c]{@{}l@{}}1. Kernel area not provided. \\ 2. Kernel perimeter information missing. \\ 3. Kernel compactness data not recorded. \\ 4. Length of kernel data missing. \\ 5. Width of kernel data missing. \\ 6. Asymmetry coefficient information missing. \\ 7. Length of kernel groove information missing.\end{tabular}                                                                                                                                                                                                                                                                                                                                                                                                                                                                                                      \\ \hline
\end{tabular}}
\end{table*}

\subsection{Optimal $k$ Values for k-NN Imputation in Various Missingness Patterns}
\label{appendix:optimal-k-values}

Table \ref{tab:optimal_k} shows the optimal values of $k$ for k-NN imputation across three missingness patterns (MCAR, MAR, and MNAR) using the Llama and FLAN-T5 models on six datasets. These optimal values were determined through hyperparameter tuning, where k was varied between 3 and 9, based on the downstream classification task to achieve the best imputation performance for each dataset and missingness pattern combination. This tuning process ensures that the k-NN imputation method is tailored to the specific characteristics and requirements of each dataset, enhancing overall performance.

\begin{table}[htb!]
\centering
\caption{Optimal $k$ values for k-NN imputation across MCAR, MAR, and MNAR missingness patterns using Llama and FLAN-T5 models on six datasets.}
\label{tab:optimal_k}
{\small
 \begin{tabular}{p{1.6cm} p{1.2cm} p{1.2cm} p{0.3cm} p{1.3cm} }
\toprule
\textbf{Dataset} & \textbf{Missing} & \textbf{Model} & \textbf{$k$} & \textbf{Accuracy} \\
 & \textbf{pattern}  & &  &  \textbf{(\%)}  \\
\midrule
\multirow{6}{*}{\textbf{Iris}} & \multirow{2}{*}{MCAR} & Llama & 5 & 84.60 \\
 &  & FLAN-T5 & 5 & 78.80 \\
 & \multirow{2}{*}{MAR} & Llama & 5 & 87.40 \\
 &  & FLAN-T5 & 3 & 82.20 \\
 & \multirow{2}{*}{MNAR} & Llama & 7 & 76.60 \\
 &  & FLAN-T5 & 5 & 72.20 \\
\midrule
\multirow{6}{*}{\textbf{Wine}} & \multirow{2}{*}{MCAR} & Llama & 3 & 80.20 \\
 &  & FLAN-T5 & 3 & 74.60 \\
 & \multirow{2}{*}{MAR} & Llama & 5 & 86.20 \\
 &  & FLAN-T5 & 5 & 82.40 \\
 & \multirow{2}{*}{MNAR} & Llama & 5 & 73.20 \\
 &  & FLAN-T5 & 3 & 71.60 \\
\midrule
\multirow{6}{*}{\textbf{Seeds}} & \multirow{2}{*}{MCAR} & Llama & 3 & 79.40 \\
 &  & FLAN-T5 & 3 & 79.00 \\
 & \multirow{2}{*}{MAR} & Llama & 3 & 81.60 \\
 &  & FLAN-T5 & 5 & 81.20 \\
 & \multirow{2}{*}{MNAR} & Llama & 3 & 72.20 \\
 &  & FLAN-T5 & 5 & 71.60 \\
\midrule
\multirow{6}{*}{\textbf{\begin{tabular}{@{}c@{}}Glass\end{tabular}}} & \multirow{2}{*}{MCAR} & Llama & 5 & 52.40 \\
 &  & FLAN-T5 & 5 & 44.20 \\
 & \multirow{2}{*}{MAR} & Llama & 5 & 57.60 \\
 &  & FLAN-T5 & 5 & 49.20 \\
 & \multirow{2}{*}{MNAR} & Llama & 3 & 41.40 \\
 &  & FLAN-T5 & 5 & 36.60 \\
\midrule
\multirow{6}{*}{\textbf{Ionosphere}} & \multirow{2}{*}{MCAR} & Llama & 5 & 86.80 \\
 &  & FLAN-T5 & 5 & 87.20 \\
 & \multirow{2}{*}{MAR} & Llama & 5 & 85.80 \\
 &  & FLAN-T5 & 5 & 83.40 \\
 & \multirow{2}{*}{MNAR} & Llama & 3 & 79.60 \\
 &  & FLAN-T5 & 5 & 78.20 \\
\midrule
\multirow{6}{*}{\textbf{Cancer}} & \multirow{2}{*}{MCAR} & Llama & 5 & 85.20 \\
 &  & FLAN-T5 & 3 & 83.00 \\
 & \multirow{2}{*}{MAR} & Llama & 5 & 89.80 \\
 &  & FLAN-T5 & 5 & 85.60 \\
 & \multirow{2}{*}{MNAR} & Llama & 5 & 82.40 \\
 &  & FLAN-T5 & 5 & 78.40 \\
\bottomrule
\end{tabular}
}
\end{table}

\end{document}